\documentclass[sigconf]{acmart}
\usepackage{enumitem}
\usepackage{multirow}
\usepackage{subfigure}
\usepackage{algorithm}
\usepackage{algorithmic}
\usepackage{xspace}
\usepackage{tabularx}

\AtBeginDocument{%
  \providecommand\BibTeX{{%
    \normalfont B\kern-0.5em{\scshape i\kern-0.25em b}\kern-0.8em\TeX}}}

\setcopyright{acmcopyright}
\copyrightyear{2024}
\acmYear{2024}
\acmDOI{XXXXXXX.XXXXXXX}

\acmPrice{15.00}
\acmISBN{978-1-4503-XXXX-X/18/06}

\newcommand{\method}{{MITIGATE}\xspace}

\begin{document}

\title{Multitask Active Learning for Graph Anomaly Detection}

\author{Wenjing Chang}
\affiliation{%
  \institution{CNIC, CAS \and UCAS}
  \city{Beijing}
  \country{China}
}
\email{changwenjing@cnic.cn}
\author{Kay Liu}
\affiliation{%
  \institution{University of Illinois Chicago}
  \streetaddress{801 S Morgan St}
  \city{Chicago}
  \state{Illinois}
  \country{USA}
  \postcode{60607}
}
\email{zliu234@uic.edu}
\author{Kaize Ding}
\affiliation{%
  \institution{Northwestern University}
  \streetaddress{2006 Sheridan Road}
  \city{Evanston}
  \state{Illinois}
  \country{USA}
  \postcode{60208}
}
\email{kaize.ding@northwestern.edu}
\author{Philip S. Yu}
\affiliation{%
  \institution{University of Illinois Chicago}
  \streetaddress{801 S Morgan St}
  \city{Chicago}
  \state{Illinois}
  \country{USA}
  \postcode{60607}
}
\email{psyu@uic.edu}
\author{Jianjun Yu}
\affiliation{%
  \institution{CNIC, CAS}
  \city{Beijing}
  \country{China}
}
\email{yujj@cnic.ac.cn}

\renewcommand{\shortauthors}{Trovato and Tobin, et al.}

\begin{abstract}
  In the web era, graph machine learning has been widely used on ubiquitous graph-structured data.
  As a pivotal component for bolstering web security and enhancing the robustness of graph-based applications, the significance of graph anomaly detection is continually increasing.
  While Graph Neural Networks (GNNs) have demonstrated efficacy in supervised and semi-supervised graph anomaly detection, their performance is contingent upon the availability of sufficient ground truth labels.
  The labor-intensive nature of identifying anomalies from complex graph structures poses a significant challenge in real-world applications.
  Despite that, the indirect supervision signals from other tasks (e.g., node classification) are relatively abundant.
  In this paper, we propose a novel \textbf{M}ult\textbf{I}task ac\textbf{TI}ve \textbf{G}raph \textbf{A}nomaly de\textbf{TE}ction framework, namely \method.
  Firstly, by coupling node classification tasks, \method obtains the capability to detect out-of-distribution nodes without known anomalies.
  Secondly, \method quantifies the informativeness of nodes by the confidence difference across tasks, allowing samples with conflicting predictions to provide informative yet not excessively challenging information for subsequent training.
  Finally, to enhance the likelihood of selecting representative nodes that are distant from known patterns, \method adopts a masked aggregation mechanism for distance measurement, considering both inherent features of nodes and current labeled status.
  Empirical studies on four datasets demonstrate that \method significantly outperforms the state-of-the-art methods for anomaly detection. Our code is publicly available at: \url{https://github.com/AhaChang/MITIGATE}.
\end{abstract}



\maketitle

\section{Introduction}

In light of the proliferation of the World Wide Web, graph-structured data has become increasingly pervasive. 
Concurrently, graph machine learning techniques have been extensively employed in various web mining tasks, such as recommendation systems~\cite{yang2021consisrec}, community detection~\cite{jin2021survey}, traffic forecasting~\cite{zhang2021graph}, etc.
To ensure the robustness and security of such graph learning-based applications in web environments, graph anomaly detection serves as an indispensable component. 
Graph anomaly detection aims to identify abnormal substructures (e.g., nodes) in graphs that exhibit significant deviations from established norms.
It finds extensive applications in capturing high-risk entities and behaviors, including but not limited to spam detection \cite{noekhah2020opinion}, financial fraud detection \cite{wang2019semi}, and fake news detection \cite{dou2021user}.

In accordance with the insights presented in \cite{liu2022bond,ding2021few}, unsupervised methods heavily rely on the underlying data distribution to derive outlier patterns.
Consequently, these methods may exhibit unstable performance when faced with data that contains specific domain knowledge or deviates from the assumed distribution. 
However, the intricate nature of graph structures, along with the high cost of manual annotation for both normal and anomalous nodes, prevents the collection of abundant ground truth labels, thereby limiting the feasibility of applying fully supervised learning approaches.
This contradiction necessitates the exploration of alternative learning paradigms that can efficiently leverage limited supervision signals while also accommodating the complexities inherent in graph data.

Given the considerable expense of acquiring ground-truth labels for anomaly detection, it is a judicious choice to leverage the existing relatively abundant availability of labels for other graph learning tasks.
In the application of graph learning, anomaly detection can enhance the stability of various tasks (e.g., node classification) by filtering out anomalies.
Conversely, these tasks can serve as auxiliary tasks for anomaly detection and reciprocally provide external information (e.g., classification uncertainty) for augmenting the efficacy of anomaly detection. 
Besides, these auxiliary tasks inherently contain general information that can be mutually leveraged, providing an indirect supervision signals when the anomaly detection task is deficient in annotation~\cite{collobert2008unified,zhang2021survey}.

Furthermore, to effectively leverage limited direct supervision signals for anomaly detection, some semi-supervised methods have been proposed \cite{wang2019semi,dou2020enhancing,tang2022rethinking}. These methods aim to enhance the learning of anomalous patterns based on the known anomalous nodes.
The underlying assumption of these methods is that a subset of nodes including both normal and anomalous nodes have been annotated for training, and the labeled data is overall balanced. 
Nevertheless, it is non-trivial to acquire such an ideal training set from an unlabeled graph, one that contains sufficient knowledge for distinguishing normal and anomalous nodes, especially when constrained by a limited labeling budget.
Active Learning (AL) paves a promising way for addressing the labeling problem, as it enables models to enhance their learning efficiency by actively requesting the labels in training data~\cite{campbell2000query,settles2009active,aggarwal2014active,zhang2021alg}.
It has also been applied in anomaly detection tasks \cite{gornitz2013toward,ghasemi2011active,das2016incorporating}, aiming at discovering more anomalies based on heuristic query strategies, such as uncertainty-based, diversity-based, and anomaly score-based strategies. 
In this way, a selection of samples that is more likely to contain a relatively higher proportion of anomalies can be used to fine-tune the model iteratively.
However, existing query strategies for anomaly detection are primarily designed for independent and identically distributed data, which hardly consider relationships among samples and may not be well-suited for graph-structured data.
Therefore, we urgently need a query strategy specifically for graph data to provide powerful direct supervision signals for anomaly detection.

To leverage the direct and indirect supervision signals efficiently and effectively, we propose a \textbf{M}ult\textbf{I}task ac\textbf{TI}ve \textbf{G}raph \textbf{A}nomaly de\textbf{TE}ction framework (\method). It incorporates external supervision signals from auxiliary tasks and productively exploits direct supervision signals by actively labeling nodes for anomaly detection.
Specifically, we first consider a node classification task together with the anomaly detection task. 
We initialize the multitask framework by node classification and detect out-of-distribution samples (i.e., anomalies) with classification uncertainty.
To query more valuable nodes, we then introduce a dynamic informativeness metric that relies on confidence difference and a representativeness metric based on the masked aggregation mechanism.
Initially, in the absence of anomalies, we prioritize the informativeness metric to focus more on the classification uncertainty for picking out-of-distribution nodes to label. 
Afterward, in order to mitigate the variation in predictions for the same node across the two tasks, we consider nodes with greater confidence differences as being more informative.
Note that these nodes are not particularly challenging samples, given that at least one of the tasks can correctly identify them.
To enhance the diversity of selected nodes and handle the influence of neighbors that have already been labeled for selection, we re-aggregate the intermediate embedding by masking the feature of labeled neighbors and keeping a count of them. 
The former filters the previously labeled representative features, while the latter reduces the overall neighborhood information based on labeled status, thereby emphasizing the feature of the central nodes.
The main contributions of this paper are summarized as follows:
\begin{itemize}
    \item We propose \method, a novel multitask active graph anomaly detection framework to detect anomalies within a limited labeling budget, which actively queries nodes with the guidance of external supervision signals.
    \item To query more valuable nodes, we devise a dynamic strategy to measure the informativeness and representativeness of nodes according to the training and labeling status.
    \item We conduct comprehensive experiments on four datasets to verify the effectiveness of the proposed method.
\end{itemize}

\section{Problem Definition}
In this section, we formulate the problem of active learning for graph anomaly detection. 

Let $\mathcal{G} = ( \mathcal{V}, \mathbf{A}, \mathbf{X})$ denotes an attributed graph, where $\mathcal{V}=\left \{ v_1, v_2, \cdots, v_n  \right \}$ is the set of nodes, $\mathbf{A} \in {\{0,1\}}^{n \times n}$ is the adjacency matrix and $\mathbf{X} \in \mathbb{R}^{n \times k}$ is the node attribute matrix. 
Note that anomaly labels are rare in the real world, but a portion of class labels are readily accessible. 
We denote the set of nodes labeled for classification as $\mathcal{V}_{L}^{N}$, and their corresponding labels are denoted by $\mathbf{Y}^{N}_{L} \in \mathbb{R}^{n \times C}$, with $C$ representing the number of classes. $\mathbf{y}_{i}^{N}$ is the one-hot label of $v_i$. 
$\mathbf{Y}_{L}^{t} \in \{0,1\}$ is the labels for anomaly detection at the $t$-th iteration, i.e., normal or anomalous. $y_i^{A}$ is the labels of $v_i$ for anomaly detection.
We initialize a set of nodes with classification labels, i.e., $\mathcal{V}_{L}^{0}=\mathcal{V}_{L}^{N}$, and regard them as normal nodes, $\mathbf{Y}_{L}^{0}=\{0\}$.
The key notations are summarized in Appendix~\ref{app:notations}.

Given an attributed graph $\mathcal{G}$, a query strategy $\mathcal{Q}$, labeling budget $\mathcal{B}$, the goal of an AL-based anomaly detection algorithm is to select a subset of nodes denoted as $\mathcal{S}^{t}$ from the unlabeled node set $\mathcal{V}_{U}^{t-1} $, and label them in a way that minimizes the loss of the model $\mathcal{M}$:
\begin{equation}
    \underset{\mathcal{V}_{L}^{t}}{\mathrm{min}} \ \mathcal{L}(\mathcal{M},\mathcal{Q}|\mathcal{G},\mathbf{Y}^{t}_{L},\mathbf{Y}^{N}_{L}) ,
\end{equation}
where $\mathcal{V}_{L}^{t} = \mathcal{V}_{L}^{t-1} \cup \mathcal{S}^t$ and $\mathcal{V}_{U}^t = \mathcal{V}_{U}^{t-1} \setminus  \mathcal{S}^t $ are the labeled and unlabeled sets after the $t$-th selection.  $t\in \{ 1,2,...,\mathcal{B}/b \}$ and $b$ is the budget in each iteration.
Then, $\mathcal{G}$, $\mathbf{Y}^{N}_{L}$, and labels ${y}_i^{A}$ for $v_i\in \mathcal{V}_{L}^{t}$ are used to train the model $\mathcal{M}$.
For convenience, we define the labeling budget as the maximum number of nodes allowed to be labeled.

\begin{figure*}
    \centering
    \includegraphics[width=0.86\linewidth]{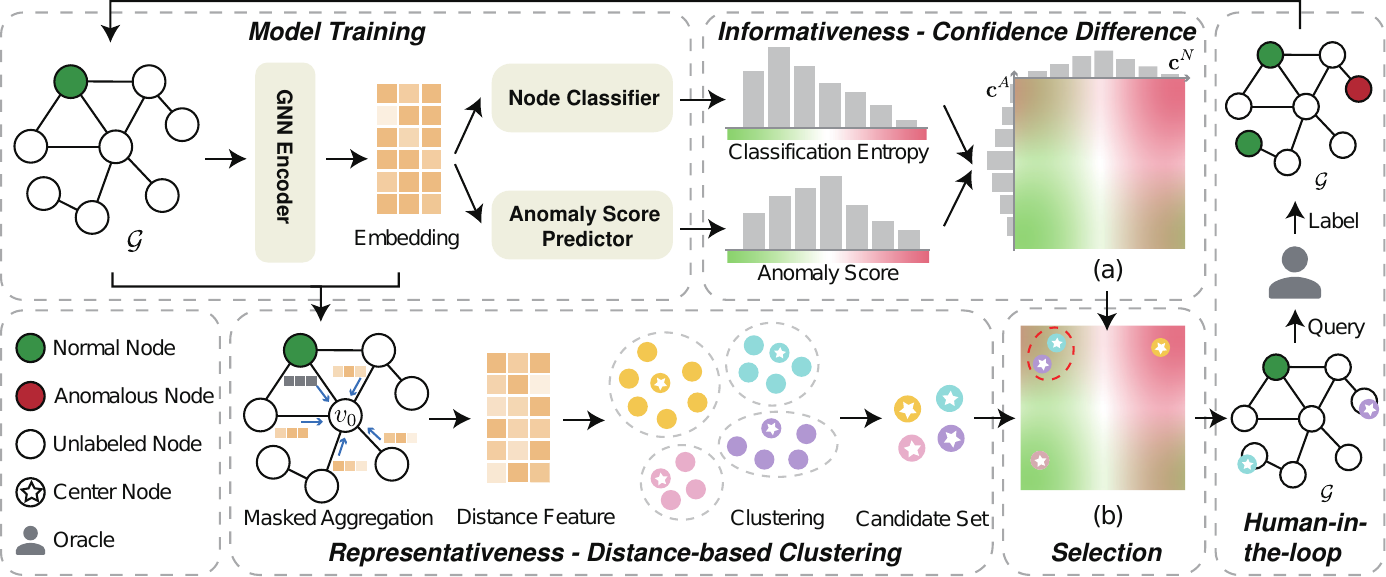}
    \caption{An illustration of the proposed \method. For a graph $\mathcal{G}$ with partial classification labels, \method employs a GNN encoder to generate node representations $\mathbf{H}$, and then employs a node classifier and an anomaly score predictor. In each selection iteration, \method assesses the representativeness and informativeness of nodes using a distance-based clustering and confidence difference across tasks for anomaly detection, respectively. Then, it picks $b$ nodes from clustering centers with high informative scores and queries an oracle to identify whether they are anomalies or not. Finally, the queried set will be incorporated into the labeled set, and continue training of the model.
    (a) A 2-dimensional confidence difference space. 
    (b) An example of candidates' confidence difference. We select the candidates with high confidence difference (i.e., in the upper left corner and lower right corner).}
    \label{fig:workflow}
\end{figure*}

\section{Method}
In this section, we introduce the proposed \method framework.
Firstly, we give an overview of the whole framework.
Then, we elaborate on the selection strategy including the calculation of the distance features for clustering and confidence difference across tasks in \autoref{subsec:selection}.
Finally, we introduce the training process of \method in detail in \autoref{subsec:training}.
\subsection{Overview}
\autoref{fig:workflow} illustrates the pipeline of \method.
It utilizes a shared encoder for node representation learning and two decoders for node classification and anomaly score prediction, respectively.
Considering the multitask structure, we devise a node informativeness metric based on the confidence difference across tasks. To reduce the initial performance gap, we incorporate classification uncertainty into the informativeness score measurement.
To promote diversity in node selection at each step, we employ the K-Medoids algorithm, which treats cluster centers as representative samples, with a novel distance measurement based on masked aggregation.
From these centers, we select $b$ nodes with the highest informativeness scores. 
We then provide the set of selected nodes to an oracle and obtain their labels (e.g., normal or anomalous) for the anomaly detection task.
Finally, we combine the selected set with the training set and continue training the model.
The details of the encoder, node classifier, and anomaly score predictor are as follows:
\subsubsection{Encoder}
Due to both the node classifier and anomaly score predictor needing an encoder to reflect the graph topology and node attributes into a latent space,
we adopt graph convolutional networks (GCNs) \cite{kipf2016semi} to learn expressive node representations, and the layer-wise propagation is defined as:
\begin{equation} \label{eq:h}
    \mathbf{H}^{(l+1)} = \sigma ( \tilde{\mathbf{D}}^{-\frac{1}{2}}\tilde{\mathbf{A}}\tilde{\mathbf{D}}^{-\frac{1}{2}} \mathbf{H}^{(l)} \mathbf{W}^{(l)}),
\end{equation}
where $\tilde{\mathbf{A}}=\mathbf{A}+\mathbf{I}$ and $\tilde{\mathbf{D}}_{ii}=\sum_{j}\tilde{\mathbf{A}}_{ij}$, $\mathbf{I}$ is the identity matrix and $\mathbf{W}^{(l)}$ is the weight matrix at the $l$-layer.
\subsubsection{Node classifier}
We use a graph convolutional layer as the node classifier to further preserve the structural information of intermediate node representations as follows:
\begin{equation}
    \mathbf{Z}=\sigma ( \tilde{\mathbf{D}}^{-\frac{1}{2}}\tilde{\mathbf{A}}\tilde{\mathbf{D}}^{-\frac{1}{2}} \mathbf{H} \mathbf{W}^{N}),
\end{equation}
where $\mathbf{H}$ is the final output of the encoder and $\mathbf{W}^{N}$ is the weight matrix for the node classifier. 
Due to anomalies being more likely to be uncertain in classification, we use entropy as the measure of anomaly probability, where a higher entropy score indicates a greater probability of being anomalous. The anomaly score of $v_i$ from the node classifier is
\begin{equation} \label{eq:ent}
    e_{i} = - \sum_{j}^{C} \mathbf{z}_{ij} \cdot {\log \mathbf{z}_{ij} }.
\end{equation}

\subsubsection{Anomaly score predictor}
The anomaly score predictor is built with a linear transformation along with a sigmoid function based on shared node representations:
\begin{equation}
    \mathbf{p} = \mathrm{Sigmoid} (\mathbf{W}^{A} \mathbf{H} + b^{A}),
\end{equation}
where $\mathbf{p} \in \mathbb{R}^{n}$ is the predicted anomaly scores, $W^A \in \mathbb{R}^{1\times n}$ is the weight matrix, and $b^{A} \in \mathbb{R}$ is corresponding bias term. 
\subsubsection{Hybrid anomaly score}
Given that both the node classifier and anomaly score predictor possess the ability to detect anomalies, we adopt a weighted score function to combine the standard scores of two predictions as follows:
\begin{equation} \label{eq:ascore}
    \mathbf{s} = Norm(\mathbf{e}) + \phi \cdot Norm(\mathbf{p}) ,
\end{equation}
where $\mathbf{s} \in \mathbb{R}^{n}$ is the overall anomaly score, $Norm(\mathbf{e})=\frac{\mathbf{e}-{\mu}_e} {{\sigma}_e}$, ${\mu}_e$ is the mean, ${\sigma}_e$ is the standard deviation, and $\phi$ is a hyperparameter to balance the importance of two predictions.

\subsection{Node Selection} \label{subsec:selection}
To benefit the overall performance for anomaly detection of the unified framework, we measure the value of nodes in terms of representativeness with distance-based clustering and informativeness with confidence difference. 

\subsubsection{Distance-based Clustering}
To discover representativeness samples from the huge unlabeled data pool, we devise a masked aggregation mechanism for generating distance features that consider representations in latent space and features in the previously labeled set.
Several methods \cite{cai2017active,gao2018active} adopt the Euclidean distance to measure the distance between representations $\mathbf{H}$ when performing clustering. This approach treats information equally among neighbors due to the mean aggregation mechanism in the GCN layer.
However, in the distance-based node selection, the distance features should be impacted by the current labeled status in the neighborhood. 
Specifically, the chosen nodes exhibit representational features. This is one key reason for their selection during previous iterations. Thus, directly aggregating these features may affect the representativeness of central nodes.
Therefore, we derive distance features through a masked aggregation mechanism, which considers labeled status in the neighborhood.
Initially, in order to mitigate the impact of features pertaining to labeled neighbors, their representations will be masked during the summation of neighborhood information.
Furthermore, to accentuate the distinctive features inherent to unlabeled nodes, we calculate the mean of neighborhood information according to the number of neighbors rather than the number of unlabeled neighbors. It suggests that in cases where more neighbors have been annotated, the influence of neighborhood information on the central node features will be diminished.
In the $t$-th selection, the distance features can be formulated as follows: 
\begin{equation} \label{eq:masked_h}
    \hat{\mathbf{h}}^t_i= \frac{\mathrm{SUM}( \mathbf{h}_j, \forall v_j \in \mathcal{N}(v_i) \cap \mathcal{V}_{U}^{t-1} )}{|\mathcal{N}(v_i)|}  +  \mathbf{h}_i ,
\end{equation}
where $\mathcal{N}(v_i)$ is the neighborhood of $v_i$.
Therefore, the distance between $v_i$ and $v_j$ can be calculated as follows:
\begin{equation} \label{eq:dist}
    dist(v_i,v_j) = {|| \hat{\mathbf{h}}^t_i - \hat{\mathbf{h}}^t_j ||}_2 .
\end{equation}

To this end, we combine the node features and labeled status of neighbors in the distance function, which can decrease the selection probability due to the high representative of neighbors rather than itself.
After calculating the pairwise distance, we adopt K-Medoids clustering as previous studies \cite{wu2019active,liu2022lscale}, in which centers chosen for the candidate set must be real nodes within the graph. Additionally, the number of clusters is set to $m$.
We do not focus on querying more anomalies but on learning a predictive model to effectively distinguish normal and anomalous nodes from the aspects of classification uncertainty and anomaly score.

\subsubsection{Confidence Difference}
As both the node classifier and anomaly score predictor can identify anomalies, we give a definition for the model confidence in anomaly detection.
For the node classifier, the entropy of predictions is used to justify whether a sample is anomalous as Eq.~\eqref{eq:ent}.
The higher entropy score indicates a higher confidence for a node to be classified as an anomaly. 
Also, for the anomaly score predictor, the anomaly score is the indicator to describe the level of confidence. 
The confidence of the node classifier, denoted as $\mathbf{c}^{N} \in \mathbb{R}^{n}$, and the anomaly score predictor, denoted as $\mathbf{c}^{A} \in \mathbb{R}^{n}$, are described as follows:
\begin{equation}
    \mathbf{c}^{N} \propto \mathbf{e}, \  \mathbf{c}^{A} \propto \mathbf{p} .
\end{equation}

Note that the node classifier and anomaly detector do not always perform equally on the same nodes for anomaly detection at the same stage. 
For example, a node may receive conflicting predictive discrimination from the node classifier and anomaly detector. Specifically, it has a lower value on classification entropy, which indicates it is likely to be a normal node, while it has a higher anomaly score, which means it is more likely to be anomalous. 
To eliminate the influence of scale across different tasks, 
we normalize the entropy of classification and the anomaly scores using Z-scores.
The confidence of the node classifier and anomaly score predictor can be reformulated as:
\begin{equation}
    \mathbf{c}^{N} = Norm(\mathbf{e}), \ \mathbf{c}^{A} = Norm(\mathbf{p}),
\end{equation}
where a higher value of $c_i^N$ and $c_i^A$ indicates a lower confidence of $v_i$ being normal and a higher confidence of being anomalous.
Based on this, we can employ the Manhattan distance to quantify the confidence difference as follows:
\begin{equation} \label{eq:diff}
    \mathbf{d} = {| \mathbf{c}^{A} - \mathbf{c}^{N} |}.
\end{equation}
The high confidence difference $\mathbf{d}$ indicates the controversy between two decoders. 
It's important to note that these samples with conflicting predictions in binary classification are not challenging for model training since one of the tasks can effectively handle them. 
To align the decoders and facilitate consistent prediction, human experts can provide coherent information by labeling the nodes with high confidence differences.
The visualization of confidence differences is shown in \autoref{fig:workflow}(a). 

\begin{algorithm}[t]
    \renewcommand{\algorithmicrequire}{\textbf{Input:}}
    \renewcommand{\algorithmicensure}{\textbf{Output:}}
    \caption{\method}
    \label{alg}
    \begin{algorithmic}[1]
        \REQUIRE Graph $\mathcal{G}=(\mathcal{V}, \mathbf{A}, \mathbf{X})$, query batch size $b$, total budget $\mathcal{B}$, labeled set for node classification $\mathcal{V}^{L}_{N}$, number of clusters $m$
        \ENSURE Anomaly scores $\mathbf{s}$
        \STATE $\mathcal{V}^{0}_{L}=\mathcal{V}^{N}_{L}$, $\mathcal{V}_{U}^{0}=\mathcal{V} \setminus \mathcal{V}^{N}_{L} $;
        \FOR{$t=1,2,...,\mathcal{B}/b$}
        \STATE $\mathcal{M}=\mathrm{train}(\mathcal{G},\mathcal{V}^{t-1}_{L},\mathcal{V}^{L}_{N})$;
        \STATE Calculate distance features $\hat{\mathbf{H}}^t$ with masked agg. by Eq.~\eqref{eq:masked_h};
        \STATE Calculate $dist(v_i,v_j)$ for each node in unlabeled set $\mathcal{V}^{t-1}_{U}$;
        \STATE Cluster $\mathcal{V}^{t-1}_{U}$ by K-Medoids algorithm with $m$ clusters;
        \STATE Calculate confidence difference $\mathbf{d}$ and informativeness scores $Info$ by Eq.~\eqref{eq:diff} and Eq.~\eqref{eq:info_score};
        \STATE Select the top $b$ clustering centers as $\mathcal{S}^{t}$ by $Info$;
        \STATE Query an Oracle to obtain labels for $\mathcal{S}^{t}$;
        \STATE $\mathcal{V}^{t}_{L}=\mathcal{V}^{t-1}_{L} \cup \mathcal{S}^{t}$, $\mathcal{V}^{t}_{U}=\mathcal{V}^{t-1}_{U} \setminus \mathcal{S}^{t}$;
        \ENDFOR
        \STATE $\mathcal{M}=\mathrm{train}(\mathcal{G},\mathcal{V}^{\mathcal{B}/b}_{L},\mathcal{V}^{L}_{N})$;
        \STATE Calculate the overall anomaly scores $\mathbf{s}$ by Eq.~\eqref{eq:ascore};
    \end{algorithmic}
\end{algorithm}

\subsubsection{Selection}
To judiciously select samples suitable for model training and mitigate the influence of the absence of positive samples, we introduce a time-sensitive informativeness measurement.
Firstly, as each node can exclusively belong to either the normal or anomalous category, the nodes predicted to have conflicting labels are more likely to contain crucial information essential for one of the tasks. 
Moreover, due to the scarcity of positive class samples during the initial training stages, anomalous patterns serve as more informative elements for the unified framework.
Importantly, even with a subset of class labels, the node classifier can provide an initial prediction for anomalies, which is essential for the anomaly score predictor.
We adopt an exponentially decaying parameter $\tau$ to dynamically combine the entropy scores of node classification and confidence difference across tasks during selection~\cite{liu2022lscale}.
The informativeness score $Info \in \mathbb{R}^{n} $ is defined as follows:
\begin{equation} \label{eq:info_score}
    Info = \tau ^{|\mathcal{V}_{L}^{t}|} {Norm}(\mathbf{e}) + ( 1 - \tau ^{|\mathcal{V}_{L}^{t}|} ) {\mathbf{d}} ,
\end{equation}
where $|\mathcal{V}_{L}^{t}|$ is the number of selected nodes in the $t$-th iteration, and $\tau$ can be set as a number close to 1.0, e.g., 0.99.
In all, the informativeness score is initially influenced more by anomalies, and as training progresses, it shifts the emphasis toward identifying nodes with prediction conflicts from a model-centric perspective.

Recognizing that both the informativeness and representativeness indicate the value of a node, during each iteration, \method first utilizes the aforementioned distance-based clustering algorithm to choose a subset of high representative nodes, denoted as the candidate set $\mathcal{V}_{C}^{t}$. Subsequently, a set of $b$ nodes $\mathcal{S}^t$ are chosen from $\mathcal{V}_{C}^{t}$ according to their informativeness score.
Algorithm~\ref{alg} describes the selection process together with the model training.

\subsection{Model Training} \label{subsec:training}
After each iteration of querying, \method is trained continually by optimizing from three aspects.
First, we calculate the cross-entropy loss on the pre-labeled nodes $\mathcal{V}_{L}^{N}$ for node classification, which will not be affected by selection. 
\begin{equation}
    \mathcal{L}_{nc} = -\frac{1}{|\mathcal{V}^{N}_{L}|} \sum_{v_i\in \mathcal{V}^{N}_{L}} \sum_{j=0}^{C} y^{N}_{ij} \log z_{ij} .
\end{equation}
where $\mathbf{y}^{N}_i$ is the one-hot label of node $v_i$.
Next, we employ a weighted binary cross-entropy loss, the widely used supervised loss for imbalanced data, on the labeled set $\mathcal{V}_{L}^{t}$ for anomaly detection at each iteration.
\begin{equation}
    \mathcal{L}_{ad}= -\frac{1}{|\mathcal{V}^{t}_{L}|} \sum_{v_i\in \mathcal{V}^{t}_{L}} (\gamma ^{t} y^A_i \log p_i + (1-y^A_i) \log (1-p_i) ) ,
\end{equation}
where $\gamma ^{t}$ is the ratio of anomaly to normal nodes in $\mathcal{V}^{t}_{L}$.

Note that the node classifier and anomaly score predictor rely on different types of annotations, i.e., the node classifier needs class information while the anomaly score predictor only needs to know whether a node is an anomaly or not. 
To leverage the information for the classification task from the queried set, in which nodes are only annotated as normal nodes or anomalies, we optimize the uncertainty of classification predictions on the whole labeled set, which can be formulated as:
\begin{equation} \label{eq:loss_un}
    \mathcal{L}_{un} = - \frac{1}{|\mathcal{V}^{t}_{LN}|} \sum_{v_i\in \mathcal{V}^{t}_{LN}} \sum_{k}^{C} z_{ik} \log z_{ik}
        + \frac{1}{|\mathcal{V}^{t}_{LA}|} \sum_{v_j\in \mathcal{V}^{t}_{LA}}  \sum_{k}^{C} z_{jk} \log z_{jk} ,
\end{equation}
where $\mathcal{V}^{t}_{LN}$ and $\mathcal{V}^{t}_{LA}$ denotes the normal and anomalous node set at the $t$-th iteration. In Eq.~\eqref{eq:loss_un}, we minimize the predicted uncertainty for normal nodes (the first term), while maximizing that for anomalies (the second term).

In all, the overall loss function of \method can be formulated as Eq.~\eqref{eq:loss}, where $\alpha$ and $\beta$ are weighting parameters.
\begin{equation} \label{eq:loss}
    \mathcal{L} =  \alpha \cdot \mathcal{L}_{nc} + \beta \cdot \mathcal{L}_{ad} + \mathcal{L}_{un} .
\end{equation}

\section{Experiments}
In this section, we extensively compare \method with state-of-the-art methods for anomaly detection.

\subsection{Experiments Settings}

\subsubsection{Datasets}
In our experiments, we adopt four widely used datasets with ground-truth labels for node classification. As there is no ground truth of anomaly detection, we inject contextual and structural anomalies following the previous studies \cite{ding2019deep,luo2022comga,liu2022bond} to evaluate the effectiveness for anomaly detection of our method.
We use two citation networks, Cora and Citeseer \cite{sen2008collective}, and two social networks, BlogCatalog and Flickr \cite{tang2009relational}.
The details of the datasets are described in Appendix~\ref{app:dataset}.
For each dataset, we randomly sample 500 nodes as a validation set and 1000 nodes as a test set and fix them for all methods for a fair comparison. 

\subsubsection{Baselines} \label{sec:baselines}
We compare \method with three types of baselines, including \textbf{(1) out-of-distribution (OOD) detection methods}, including GCN-ENT~\cite{kipf2016semi}, GKDE~\cite{zhao2020uncertainty}, OODGAT-ENT and OODGAT-ATT~\cite{song2022learning}, \textbf{(2) semi-supervised anomaly detection methods}, including FdGars~\cite{wang2019fdgars}, GeniePath~\cite{liu2019geniepath}, BWGNN~\cite{tang2022rethinking}, and DAGAD~\cite{liu2022dagad}, and \textbf{(3) active query strategy for anomaly detection}, including most positive query (GCN-Pos), positive diverse query (GCN-PosD) and diverse query~\cite{li2023deep} (GCN-Div). 
The details of these methods and their implementation are in Appendix~\ref{app:baselines}.

\begin{table*}[t]
\centering
\caption{Overall performance comparison in AUC-ROC(\%) and AUC-PR(\%) on four datasets. The best results are highlighted in bold, and the second best is underlined.}
\label{tab:overall}
\resizebox{0.96\linewidth}{!}{%
\begin{tabular}{c|cc|cc|cc|cc}
\hline
             & \multicolumn{2}{c|}{Cora} & \multicolumn{2}{c|}{Citeseer} & \multicolumn{2}{c|}{BlogCatalog} & \multicolumn{2}{c}{Flickr} \\ \hline
Method       & AUC-ROC     & AUC-PR      & AUC-ROC     & AUC-PR      & AUC-ROC     & AUC-PR      & AUC-ROC     & AUC-PR       \\ \hline
GCN-ENT      & 64.82$\pm$1.07 & 10.48$\pm$0.51 & 69.36$\pm$0.92 & 9.30$\pm$0.98 & 62.57$\pm$0.91 & 11.70$\pm$1.99 & 64.62$\pm$0.77 & 9.51$\pm$1.03 \\
GKDE         & 72.25$\pm$0.85 & 15.21$\pm$1.79 & \underline{74.87$\pm$0.61} & 17.07$\pm$1.55 & 44.56$\pm$0.07 & 5.29$\pm$0.02 & 44.81$\pm$0.09 & 5.05$\pm$0.03 \\
OODGAT-ENT   & 69.70$\pm$3.50 & 15.11$\pm$6.65 & 70.01$\pm$4.45 & 12.22$\pm$9.96 & 56.52$\pm$2.24 & 8.94$\pm$1.55 & 53.52$\pm$1.83 & 7.60$\pm$0.63 \\
OODGAT-ATT   & 50.76$\pm$4.71 & 5.33$\pm$0.73 & 55.33$\pm$4.95 & 6.71$\pm$1.63 & 49.51$\pm$4.18 & 6.13$\pm$1.17 & 51.17$\pm$1.24 & 6.46$\pm$0.93 \\ \hline
FdGars       & 58.55$\pm$6.22 & 6.51$\pm$1.22 & 54.61$\pm$9.09 & 5.12$\pm$1.55 & 51.99$\pm$3.38 & 6.01$\pm$0.24 & 57.95$\pm$6.61 & 13.97$\pm$8.81 \\
GeniePath    & 54.42$\pm$6.41 & 6.41$\pm$0.35 & 48.89$\pm$4.94 & 4.94$\pm$0.32 & 50.25$\pm$1.81 & 5.80$\pm$0.20 & 50.26$\pm$1.21 & 5.97$\pm$0.14\\
BWGNN        & 64.77$\pm$0.20 & 9.44$\pm$0.16 & 64.42$\pm$0.60 & 6.36$\pm$0.10 & 60.18$\pm$1.25 & 8.94$\pm$0.23 & 52.22$\pm$1.14 & 5.96$\pm$0.12 \\
DAGAD        & 58.30$\pm$0.30 & \textbf{19.17$\pm$4.83} & 61.52$\pm$3.21 & 14.10$\pm$1.82 & 56.52$\pm$2.89 & 9.03$\pm$0.92 & 63.22$\pm$0.33 & 11.66$\pm$4.40 \\ \hline
GCN-Pos     & 57.22$\pm$3.54 & 9.90$\pm$0.83 & 59.78$\pm$4.78 & 8.09$\pm$1.37 & 62.52$\pm$2.70 & 9.96$\pm$1.93 & 55.56$\pm$2.78 & 7.03$\pm$1.93 \\
GCN-PosD    & 62.24$\pm$2.07 & 12.00$\pm$1.11 & 61.54$\pm$0.68 & 7.54$\pm$0.21 & 62.42$\pm$1.63 & 9.79$\pm$1.77 & 57.24$\pm$4.85 & 10.39$\pm$4.86 \\
GCN-Div     & 61.23$\pm$2.88 & 9.05$\pm$1.00 & 56.54$\pm$7.19 & 10.55$\pm$3.42 & \underline{65.51$\pm$3.86} & 11.45$\pm$1.50 & 66.89$\pm$2.02 & 10.44$\pm$1.76 \\ \hline
\method-A   & \underline{73.59$\pm$3.53} & 17.00$\pm$2.55 & 71.25$\pm$3.00 & \underline{20.21$\pm$6.48} & 63.94$\pm$3.08 & \underline{12.40$\pm$2.50} & \underline{68.89$\pm$3.37} & \underline{15.69$\pm$1.72} \\
\method-E   & 72.81$\pm$1.52 & 15.98$\pm$1.95 & 74.12$\pm$1.53 & 14.60$\pm$1.97 & 63.00$\pm$2.10 & 12.09$\pm$2.67 & 68.26$\pm$3.01 & 14.18$\pm$3.83 \\
\method     & \textbf{75.45$\pm$2.27} & \underline{18.80$\pm$2.46} & \textbf{78.03$\pm$0.94} & \textbf{23.32$\pm$6.63} & \textbf{66.20$\pm$3.04} & \textbf{13.60$\pm$2.74} & \textbf{70.16$\pm$3.00} & \textbf{17.33$\pm$1.95} \\ \hline
\end{tabular}%
}
\end{table*}

\begin{figure*}
    \centering
    \includegraphics[width=0.98\linewidth]{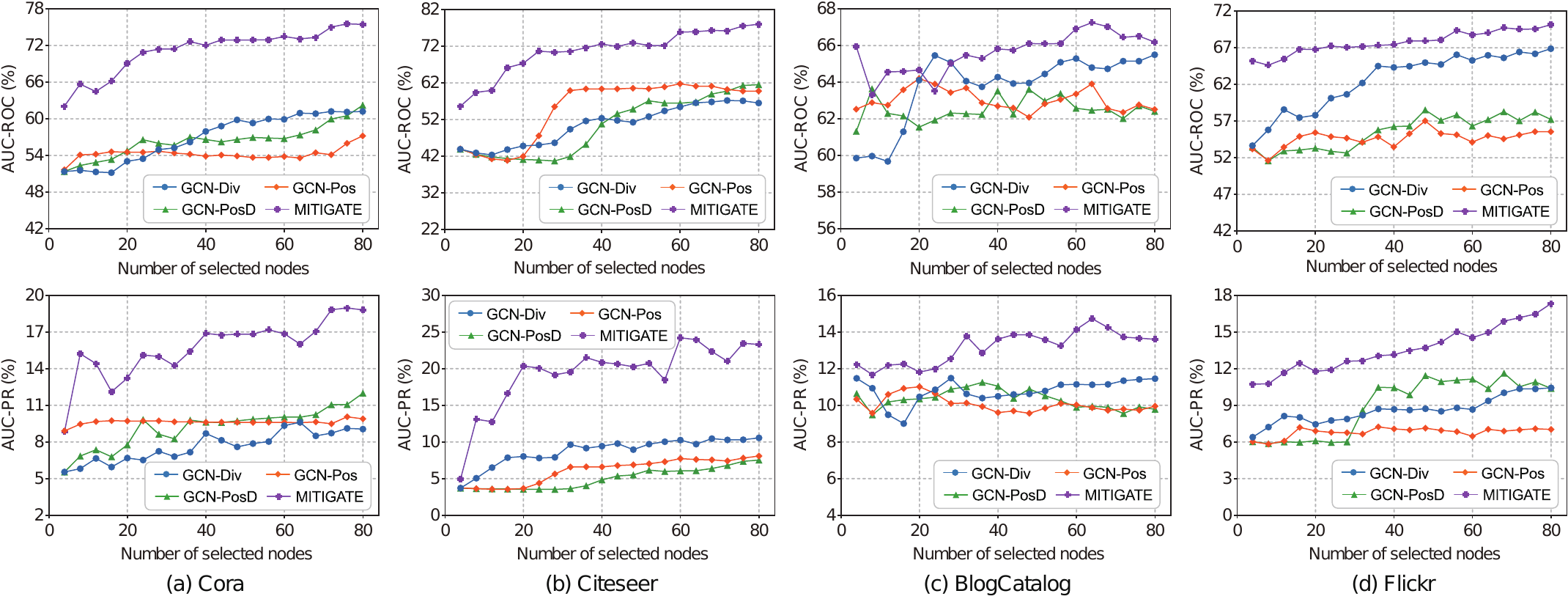} 
    \caption{Performance over different numbers of labeled nodes in selection averaged from 5 runs on four datasets.}
    \label{fig:budget}
\end{figure*}

\subsubsection{Implementation Detail}
In our experiments, we set the max budget $\mathcal{B}=80$, which means 80 nodes can be labeled, and we set $b=4$ at each iteration.  
We assume $20k$ nodes have been annotated with classification labeled, where $k$ is the number of classes, and no anomalies as initial.
For \method, we adopt Adam optimizer and the learning rate is set to 0.01. 
As to $\alpha$ and $\beta$ in the overall loss function (Eq.~\eqref{eq:loss}), $\tau$ in the informativeness score function (Eq.~\eqref{eq:info_score}), the number of clusters $m$ in K-Medoids algorithm, we tune the hyperparameters and select the best-performing results according to the validation set, the details are shown in Appendix~\ref{app:implementation}.
We set the maximum epochs of training in each iteration to 300 and perform early-stopping when ($AUCROC+Accuracy$) stops to increase for 20 epochs. $Accuracy$ is the accuracy of the node classifier on in-distribution nodes and $AUCROC$ is the performance evaluation of the anomaly score predictor.
We implement two variants of \method, namely \textbf{\method-A} and \textbf{\method-E}. \method-A uses predicted anomaly scores, while \method-E employs the entropy of classification as the final score for anomaly detection.

We adopt two widely used complementary measures in previous studies, including the Area Under Receiver Operating Characteristic Curve (AUC-ROC) and Area Under Precision-Recall Curve (AUC-PR).
To mitigate results randomness, we run the proposed method and baselines 5 times with different random seeds and record the average scores and standard deviation.

\subsection{Evaluation Results}
We evaluate the anomaly detection performance of \method and all compared methods mentioned in Section~\ref{sec:baselines} on the four datasets. The results are shown in \autoref{tab:overall} and \autoref{fig:budget}.
\subsubsection{Overall Comparison}
We evaluate the overall performance when the labeling budget $\mathcal{B}$ is set to 80 (i.e., a maximum of 80 nodes can be labeled). The corresponding AUC-ROC and AUC-PR are reported in \autoref{tab:overall}. 
We observe that \method significantly outperforms other methods under most metrics.
\textbf{(1) Comparison with classification methods.} GKDE and OODGAT are two state-of-the-art methods for OOD detection, which only utilize a portion of labeled in-distribution nodes for training. Different from them, our \method adopts anomalies (i.e., out-of-distribution nodes) in the training process to better learn a decision boundary for anomaly detection. 
Experimental results demonstrate that \method outperforms these two methods, with at least 3.2\% and 3.6\% improvement in AUC-ROC and AUC-PR, respectively.
\textbf{(2) Comparison with semi-supervised anomaly detection methods.}
FdGars, GeniePath, BWGNN, and DAGAD are superior methods for node anomaly detection. However, these methods heavily rely on labeled data.
When the labeling budget is low, which may lead to the absence of anomalies, these methods become frangible. 
It is evident that \method outperforms them, particularly in terms of AUC-ROC.
\textbf{(3) Comparison with query strategies for anomaly detection.}
These query strategies focus on selecting anomalies or high-uncertainty samples in anomaly detection. 
Though they may choose more anomalies, their performance remains suboptimal due to anomalies not always contributing significantly to the model 
and the potential challenges posed by uncertain nodes. 
More specifically, an anomalous sample, sharing similarities with known anomalous patterns, can yield a high anomaly score but may not provide substantially novel information to the model.
Besides, hard samples that contain noisy information for model training also exhibit high uncertainty, and selecting such samples potentially hinders the model performance.
Unlike them, \method selects samples based on the confidence difference across tasks.
This implies that the chosen samples are informative for at least one task and are not excessively challenging, as one of the tasks can accurately identify them.
\textbf{(4) Comparison with \method varients.} 
\method-A, \method-E, and \method differ in their final score for anomaly detection. 
It is observed that the results of using a single indicator do not differ significantly in identifying anomalies in most cases, whereas the weighted sum of these two components consistently offers discernible advantages. 
This is because \method-A and \method-E detect anomalies from a single aspect, namely the uncertainty of classification and the learnable anomaly scores, respectively, without incorporating external information from other facets.
Moreover, this observation highlights the effectiveness of integrating classification tasks and anomaly detection tasks in the process of identifying anomalies. 

\subsubsection{Performance under Different Budget}
We evaluate the performance of our proposed \method and several active query strategies for anomaly detection over the different numbers of labeled nodes for training. 
The results are shown in \autoref{fig:budget}.
Compared with the other baselines, we can observe that \method achieves the best performance in AUC-ROC under each labeling budget in most of the compared settings.
In particular, to achieve the AUC-ROC of 66.8\% on Flickr, GCN-Div labels 80 nodes while \method only labels 16 nodes. 
We contribute the effectiveness of \method at a low labeling budget with the classification task, which can provide initial discrimination for anomalies and alleviate the absence of anomalies problem.  

\begin{figure}
    \centering
    \subfigure[AUC-ROC]{\includegraphics[width=0.49\linewidth]{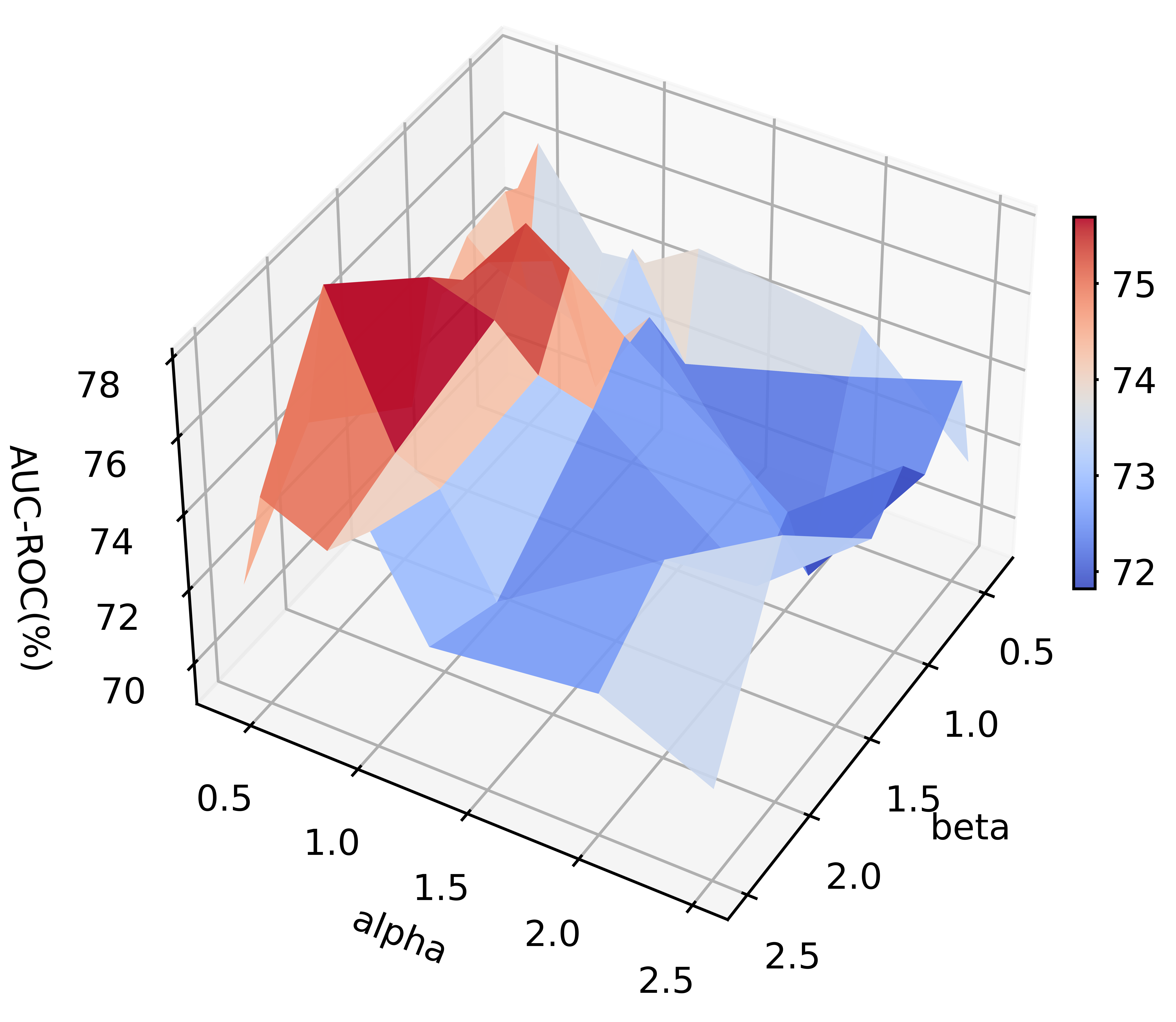}}
    \subfigure[AUC-PR]{\includegraphics[width=0.49\linewidth]{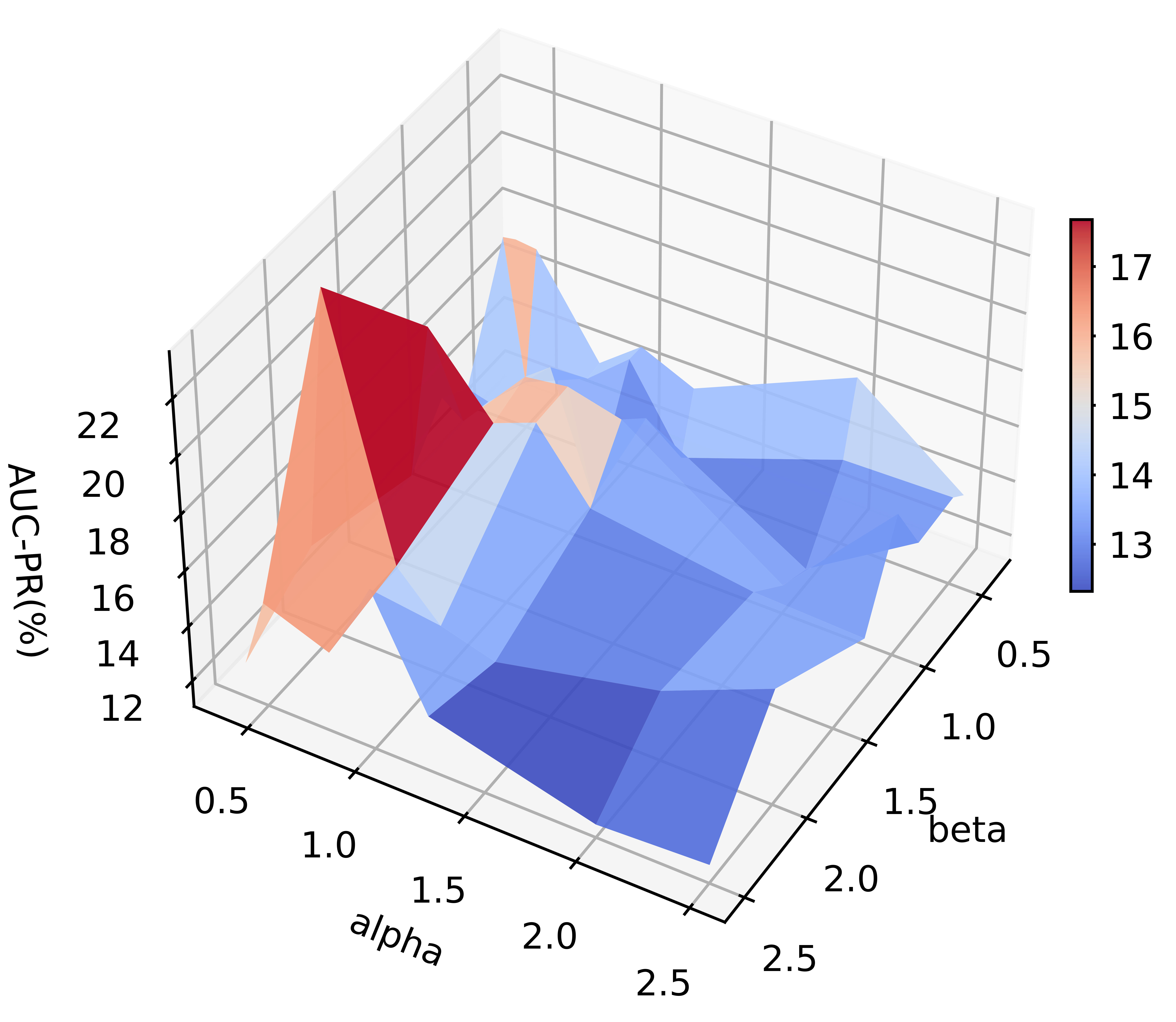}}
    \caption{Weight analysis for the node classifier and anomaly score predictor with various values of $\alpha$ and $\beta$ on Citeseer.}
    \label{fig:sensitive_alpha_beta}
    \vspace{-0.1in}
\end{figure}

\subsection{Parameter Analysis}
We further analyze the majority of parameters in \method on Citeseer.
The results are shown in \autoref{fig:sensitive_alpha_beta} and \autoref{fig:sensitive}. 
\subsubsection{Impacts of $\alpha$ and $\beta$}
We analyze the impact of varying weights in the overall loss function as shown in Eq.~\eqref{eq:loss}.
These weights are crucial in achieving a balance among the three components of the loss, namely the node classification loss, anomaly detection loss, and uncertainty loss.
We evaluate \method for $\alpha \in \{ 0.4, 0.5, 0.8, 1.0, 1.25, 2, 2.5 \}$ and $\beta \in \{0.4, 0.5, 0.8, 1.0, 1.25, 2, 2.5 \}$ and report the average results of 5 runs in \autoref{fig:sensitive_alpha_beta}.
We can observe that lower values of $\alpha$ and higher values of $\beta$ lead to better performances. 
One possible reason is that the category characteristics in Citeseer are easy to differentiate. 
It further suggests that adjusting $\alpha$ and $\beta$ properly can bring more benefits to the overall performance. 

\subsubsection{Impacts of $\phi$} 
The weight term $\phi$ is essential in calculating the final anomaly scores by balancing the importance of classification uncertainty and learnable anomaly scores in Eq.~\eqref{eq:ascore}. 
\autoref{fig:s_phi} presents the AUC-ROC and AUC-PR of \method when varying $\phi \in \{ 1.6, 1.8, 2, 2.2, 2.4 \}$.  
It is evident that although both terms are capable of identifying anomalies, their contributions are not equal. The optimal weight ratio between classification uncertainty and learnable anomaly score for the final anomaly score is 1:2 (i.e., $\phi=2$), indicating the anomaly predictor exerts a more pronounced influence in identifying anomalies.

\subsubsection{Impacts of $m$}
\begin{figure}[t]
    \centering
    \subfigure[$\phi$]{ \label{fig:s_phi}
    \includegraphics[width=0.48\linewidth]{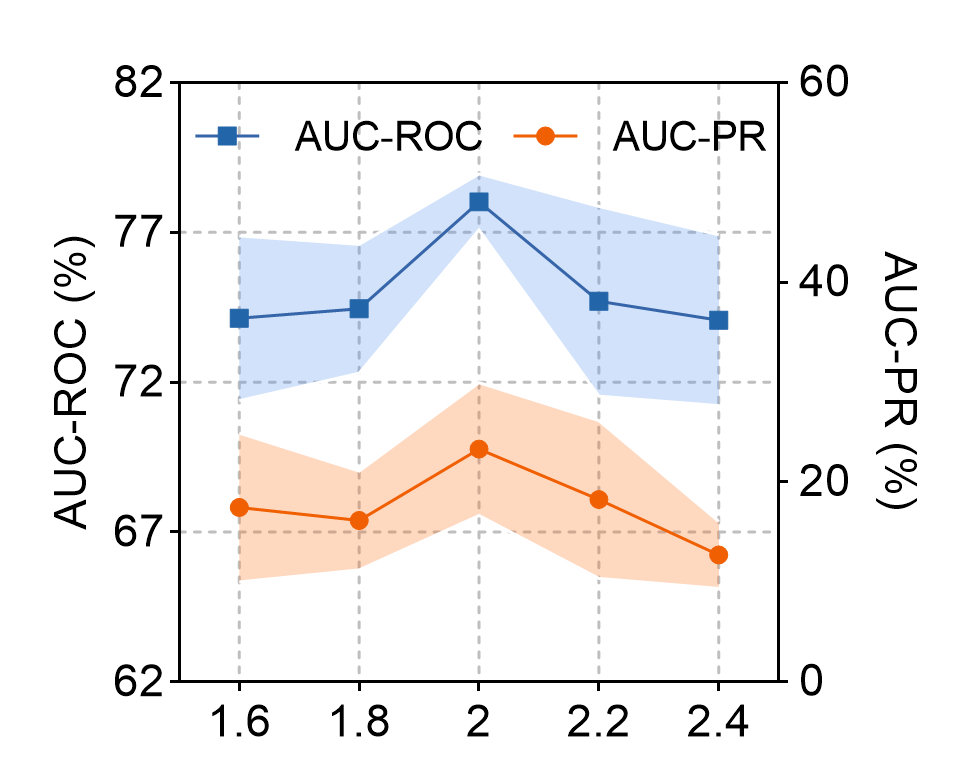}} 
    \subfigure[$m$]{ \label{fig:s_c_num}
    \includegraphics[width=0.48\linewidth]{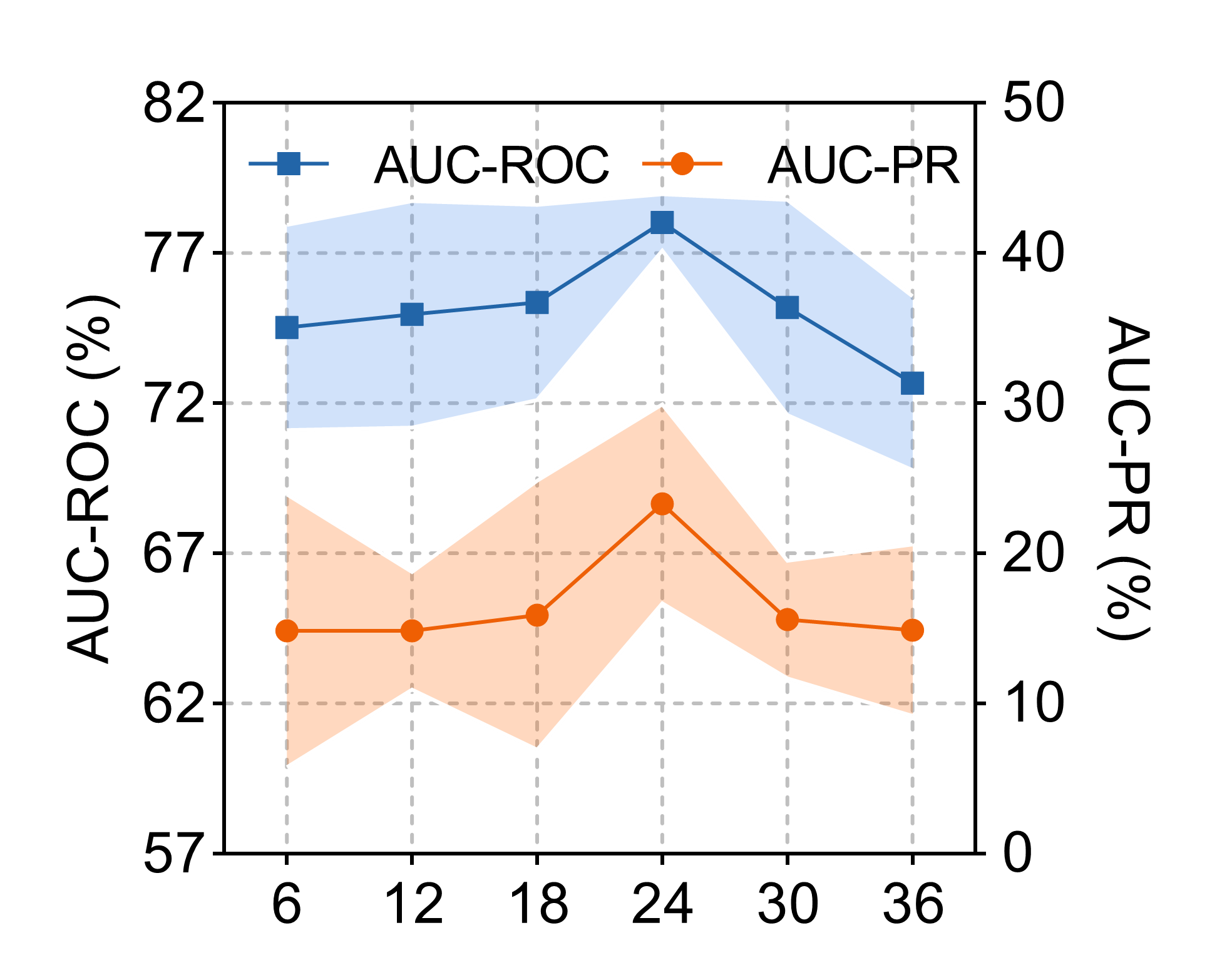}}
    \caption{Sensitivity analysis on Citeseer for (a) overall anomaly scores weight term $\phi$, and (b) number of cluster $m$.}
    \label{fig:sensitive}
\end{figure}

As clustering plays a pivotal role in choosing representative nodes from the unlabeled set, we investigate the performance of \method by varying the number of clusters $m$ from 1 to 6 times the class number. 
As shown in \autoref{fig:s_c_num}, we can observe that the optimal value of $m$ tends to be near $4k$, which leads to the optimal AUC-ROC and AUC-PR. 

\begin{table}[t]
\caption{Ablation study on Citeseer and Flickr. The best results are highlighted in bold.}
\label{tab:abaltion}
\vspace{-0.1in}
\resizebox{\columnwidth}{!}{%
\begin{tabular}{cl|cc}
\hline
Dataset                   & Variants     & AUC-ROC       & AUC-PR        \\ \hline
\multirow{6}{*}{Citeseer} & w/o uncertainty loss & 72.33$\pm$1.55 & 10.80$\pm$1.71 \\
                          & w/o entropy score & 76.35$\pm$3.77 & 15.75$\pm$4.77 \\
                          & w/o confidence difference & 73.62$\pm$1.77 & 14.79$\pm$3.85 \\
                          & w/o masked aggregation & 71.95$\pm$4.94 & 11.81$\pm$2.94 \\
                          & w/o clustering & 69.33$\pm$1.87 & 21.22$\pm$1.69 \\
                          & \method     & \textbf{78.03$\pm$0.94} & \textbf{23.32$\pm$6.63} \\ \hline
\multirow{6}{*}{Flickr}   & w/o uncertainty loss & 66.13$\pm$0.95 & 10.43$\pm$1.45 \\
                          & w/o entropy score & 65.51$\pm$3.12 & 10.06$\pm$1.82 \\
                          & w/o confidence difference & 64.85$\pm$1.62 & 12.76$\pm$1.63 \\
                          & w/o masked aggregation & 67.57$\pm$3.58 & 12.62$\pm$2.44 \\
                          & w/o clustering & 68.48$\pm$2.04 & 12.45$\pm$3.01 \\
                          & \method     & \textbf{70.16$\pm$3.00} & \textbf{17.33$\pm$1.95} \\ \hline
\end{tabular}%
}
\vspace{-0.1in}
\end{table}

\subsection{Ablation Study}
We conduct an ablation study to examine the contribution of each key component in the proposed framework on Citeseer and Flickr. The results are shown in \autoref{tab:abaltion}.
\begin{itemize}[leftmargin=*]
    \item \textbf{w/o uncertainty loss}: it removes the uncertainty loss on selected nodes in Eq.~\eqref{eq:loss}.
    \item \textbf{w/o entropy score}: it removes the classification uncertainty score in informativeness measurement in Eq.~\eqref{eq:info_score}.
    \item \textbf{w/o confidence difference}: it removes the confidence difference between tasks in informativeness measurement in Eq.~\eqref{eq:info_score}.
    \item \textbf{w/o masked aggregation}: it replaces distance features calculated through masked aggregation as Eq.~\eqref{eq:masked_h} with the representation in latent space obtained from Eq.~\eqref{eq:h}.
    \item \textbf{w/o clustering}: it removes the distance-based clustering, which aims to discover representative nodes.
\end{itemize}
\subsubsection{Uncertainty Loss}
As shown in \autoref{tab:abaltion}, we see that \method notably outperforms \method w/o uncertainty loss, exhibiting improvements of 5.7\% and 12.5\% in terms of AUC-ROC and AUC-PR, respectively.
It indicates that incorporating the uncertainty loss from the classification perspective on selected nodes, which are exclusively labeled as normal or anomalous,  can substantially improve the anomaly detection performance of \method.
\subsubsection{Strategy of Node Selection}
To verify the effectiveness of the proposed selection strategy, we conduct ablation tests on four variants.
First, compared with \method w/o entropy score and \method w/o confidence difference, which remove partial of the informativeness score respectively, \method achieves the best performance.
It indicates that both the classification uncertainty and confidence difference contribute to the node informativeness, and the dynamic combination of them is also effective.
This is expected since the uncertainty loss is not a direct supervision for the node classifier and the performance of the anomaly score predictor performs worse at the beginning without anomalous samples. 
Besides, the comparison between \method w/o masked aggregation and \method w/o clustering indicates that the clustering makes a minimal or even negative contribution without efficient distance measurement. 
For instance, \method w/o clustering surpasses \method w/o masked aggregation by approximately 10\% in terms of AUC-PR on Citeseer.
Furthermore, when comparing \method and \method w/o masked aggregation, it is obvious that the proposed masked aggregation effectively improves the discriminative capability in anomaly detection.
We argue that by masking representations for previously labeled nodes within the neighborhood, more representative nodes are selected in relation to both labeled and unlabeled sets.

\section{Related Work}
\subsection{Active Learning for Anomaly Detection}
Active learning aims to interactively select samples from unlabeled data to maximize model performance within limited labeling budgets. It has been extensively studied in the field of anomaly detection \cite{gornitz2013toward,ghasemi2011active,das2016incorporating,zha2020meta}.
In contrast to traditional active learning, active anomaly detection focuses on discovering more anomalous samples.
Several methods incorporate active queried data with the unsupervised learning paradigm in the training process. For instance, \cite{gornitz2013toward} suggests querying samples close to the decision boundary, which means more uncertainty. \cite{ghasemi2011active} suggests querying samples based on density to ensure a diverse distribution of predicted anomalous data for querying. Devising an effective query strategy to discover more anomalies becomes an important problem.
\cite{das2016incorporating,ning2022deep} aim to query the most anomalous sample according to predicted scores, while \cite{das2019active} incorporates the density information with anomaly score. 
However, these greedy strategies prioritize short-term gains and may yield suboptimal results. To address this issue, deep reinforcement learning has been introduced to the design of query strategies. For example, Meta-AAD \cite{zha2020meta} trains a meta-policy using auxiliary labeled datasets so that it can be directly applied to new unlabeled datasets without further tuning.
In our work, we focus on improving the predictive performance for graph anomaly detection by leveraging corresponding auxiliary tasks to help the model training and sample querying processes.
\subsection{Graph Anomaly Detection}
Graph anomaly detection methods \cite{wang2019fdgars,wang2019semi,tang2022rethinking,chai2022can} have been extensively studied in recent years, which assumes that a set of nodes have been labeled. 
Motivated by general GNN algorithms, several graph anomaly detection methods based on redesigned message passing and aggregation mechanisms have been proposed. 
For example, FdGars \cite{wang2019fdgars} utilizes GCN to combine the characteristics of reviewers and their relationships.
Semi-GNN \cite{wang2019semi} employs hierarchical attention to model the multi-view graph for fraud detection.
GraphConsis \cite{liu2020alleviating} proposes to filter inconsistent neighbors in aggregation to maintain the unique semantic characteristics of the target node.
Moreover, the imbalance between the majority and minority classes is another essential problem in anomaly detection.
To alleviate this problem, PC-GNN \cite{liu2021pick} incorporates label distribution information to sample neighbors in the aggregation process, while DAGAD \cite{liu2022dagad} devises a graph data augmentation module to fertilize training set with generated samples. 
Different from the aforementioned spatial methods, BWGNN \cite{tang2022rethinking} leverages spectral distribution information to capture graph anomalies with high-frequency and AMNet \cite{chai2022can} adaptively combines signals of low-frequency and high-frequency to learn the node embedding for distinguishing the anomalous nodes.
However, these methods may not work well when the labeling budget is relatively low since they highly rely on the initial labeled set, i.e., at least one negative sample needs to be labeled. 
Whereas, we focus on actively selecting nodes for annotation during the training process, which means the selected labeled nodes can bring more benefits to the target model.

\subsection{Out-of-distribution Detection}
Out-of-distribution (OOD) detection aims to distinguish samples drawn from a distribution different from the labeled in-distribution samples \cite{zhou2021step}. 
Extensive studies \cite{liang2018enhancing,yu2019unsupervised,sehwag2020ssd,zhou2021contrastive} have been proposed for OOD detection on structured data, such as text and images. 
However, graph data contains not only structured attributes but also non-Euclidean topology structures. Consequently, several graph-specific OOD detection methods have been proposed, including uncertainty-based methods \cite{zhao2020uncertainty,stadler2021graph} and graph learning-based methods \cite{song2022learning,huang2022end}. 
Uncertainty-based methods aim to use the confidence of a well-trained model on partial ID samples. GKDE \cite{zhao2020uncertainty} adopts several uncertainty estimates-based metrics for OOD detection and finds the vacuity-based model has optimal performance, while GPN~\cite{stadler2021graph} devises an input-independent Bayesian update rule to model uncertainty on predicted categorical distribution.
Besides, graph learning-based methods argue that the OOD samples can affect ID samples under the message-passing mechanism. For example,  OODGAT \cite{song2022learning} explicitly models the interaction between inliners and outliers, while LMN \cite{huang2022end} learns to mix neighbors to mitigate the influence from OOD nodes. 
However, these methods have a limitation in that they cannot effectively utilize the supervision information from OOD samples.

\section{Conclusion}
In this paper, we proposed \method, a novel multitask active learning framework for graph anomaly detection within a limited labeling budget. \method comprises a node classifier and an anomaly score predictor, with the primary goal of enhancing the overall performance of anomaly detection through the selection of the queried set.
Besides, the masked aggregation mechanism for distance features proves instrumental in selecting representative nodes.
Additionally, node informativeness is dynamically measured by the confidence difference across tasks and classification uncertainty.
Experimental results on four datasets demonstrate the effectiveness of \method compared with the state-of-the-art methods.

\bibliographystyle{abbrv}
\bibliography{main}

\appendix
\section{Notations} \label{app:notations}
Here we list the key notations in our paper in \autoref{tab:notations}.

\begin{table}[h]
\caption{Summary of notations.}
\label{tab:notations}
\begin{tabularx}{\columnwidth} {
         >{\hsize=.18\hsize\linewidth=\hsize}X
         >{\hsize=\hsize\linewidth=\hsize}X
        }
\hline
Notation & Description \\ \hline
$\mathcal{G}, \mathbf{A}, \mathbf{X}, \mathcal{V}$ & Graph, adjacency matrix, attribute matrix, node set. \\
$\mathcal{B}, b$ & The total budget, query batch size. \\
$\mathcal{V}_{L}^{N}$ & The set of nodes labeled for classification. \\
$\mathcal{V}_{L}^{t}$ & The set of labeled nodes at the $t$-th iteration. \\
$\mathcal{V}_{U}^{t}$ & The set of unlabeled nodes at the $t$-th iteration. \\
$\mathcal{S}^{t}$ & The set of selected nodes  at the $t$-th iteration. \\
$\mathbf{e}$ & The entropy of predictions from node classifier. \\
$\mathbf{p}$ & The anomaly score from anomaly score predictor. \\
$\mathbf{s}$ & The overall anomaly score. \\
$\mathbf{h}_i^t$ & The latent embedding of $v_i$ at the $t$-th iteration. \\
$\hat{\mathbf{h}}_i^t$ & The distance feature of $v_i$ at the $t$-th iteration. \\
$dist(v_i,v_j)$ & The masked distance between $v_i$ and $v_j$. \\
$\mathbf{c}^N,\mathbf{c}^A$ & The confidence of node classifier and anomaly score predictor in identifying anomalies. \\
$\mathbf{d}$ & The confidence difference. \\
$Info$  & The informativeness score. \\
$\tau$ & The decaying parameter in informativeness score. \\
$\phi$ & The weight term in overall anomaly score. \\
$\alpha$, $\beta$    & Weight of classification and anomaly detection loss. \\
$m$ & The number of clusters. \\ \hline
\end{tabularx}%
\end{table}

\section{DETAILS OF EXPERIMENTAL SETTINGS}
\subsection{Datasets} \label{app:dataset}
We employ four datasets to evaluate the performance of \method. 
Cora and Citeseer~\cite{sen2008collective} are two citation networks, in which nodes represent papers while edges represent citation relationships among papers. The class labels denote corresponding academic fields.
BlogCatalog and Flickr~\cite{tang2009relational} are two social networks, in which nodes represent users while edges represent social relationships between users. The class labels denote the interests of users. 
The dataset statistics are shown in~\autoref{tab:dataset}.

Following the previous studies \cite{ding2019deep,liu2021anomaly,liu2022bond,liu2022dagad}, we inject contextual and structural anomalies into four widely used datasets: Cora, CiteSeer, Flickr, and BlogCatalog.
\textbf{(1) Structural anomalies}. The idea behind synthetic structural anomalies is that anomalies can manifest as densely connected nodes within small cliques \cite{skillicorn2007detecting}. For every clique, $p$ nodes are randomly selected and interconnected completely. Repeating this step $q$ times to generate $q$ cliques, ultimately injecting $p \times q$ structural anomalies. In our experiments, we fix $p=15$ and set $q$ to $10, 15, 5, 5$ for BlogCatalog, Flickr, Cora, and CiteSeer, respectively.
\textbf{(2) Contextual anomalies}. The contextual anomalies are injected by perturbing the attributes of nodes. Initially, a node $v_i$ is randomly selected from the vertex set $V$. Then,  $k$ additional nodes are sampled from $V \setminus v_i$, forming a subset $V_c$. For each node $v_c \in V_c$, we measure the Euclidean distance to $v_i$ and subsequently adjust the attributes of $v_i$ to be the same as $v_c$ with the largest distance. In our experiments, we set $k=50$  and the number of contextual anomalies as $p \times q$, following previous studies to ensure an adequately large disturbance magnitude and to balance the proportions of different anomaly types \cite{liu2021anomaly}.

\begin{table}[h]
\centering
\caption{Dataset Analysis}
\label{tab:dataset}
\resizebox{\columnwidth}{!}{%
\begin{tabular}{c|ccccc}
\hline
Dataset     & \#Nodes  & \#Edges  & \#Attributes & \#Class  & \#Anomalies \\ \hline
Cora        & 2,708    & 5,429    & 1,433        & 7        & 150         \\
CiteSeer    & 3,327    & 4,732    & 3,703        & 6        & 150         \\
BlogCatalog & 5,196    & 171,743  & 8,189        & 6        & 300         \\
Flickr      & 7,575    & 239,738  & 12,047       & 9        & 450         \\ \hline
\end{tabular}%
}
\end{table}

\subsection{Baselines} \label{app:baselines}
In our experiments, the compared methods include 
(1) OOD detection methods (GCN-ENT, GKDE, OODGAT-ENT and OODGAT-ATT), 
(2) semi-supervised anomaly detection methods (FdGars, GeniePath, BWGNN and DAGAD),
(3) active query strategy for anomaly detection (most positive query, positive diverse query, and diverse query).
The details of the baselines are as follows:
\begin{itemize}[leftmargin=*]
    \item \textbf{GCN-ENT} \cite{kipf2016semi} is a vanilla GCN method that identifies anomalous nodes by uncertainty measured based on the entropy of node classification.
    \item \textbf{GKDE} \cite{zhao2020uncertainty} is a Graph-based Kernel Dirichlet GCN method for semi-supervised node classification and OOD detection.
    \item \textbf{OODGAT} \cite{song2022learning} is a graph learning method with OOD nodes to distinguish inliers from outliers during feature propagation. In particular, \textbf{OODGAT-ENT} uses the entropy of the predicted distribution as an indicator of outliers, while \textbf{OODGAT-ATT} relies on the score provided by a binary classifier.
    \item \textbf{FdGars} \cite{wang2019fdgars} is an anomaly detection method based on GCN that expresses the characteristics of reviewers and relationships between reviewers.
    \item \textbf{GeniePath} \cite{liu2019geniepath} proposes to adaptively select receptive paths for different nodes in aggregation for anomaly detection.
    \item \textbf{BWGNN} \cite{tang2022rethinking} analyzes graph anomalies in the spectral domain and leverages Beta graph wavelet to better capture anomaly information on graphs.
    \item \textbf{DAGAD} \cite{liu2022dagad} devises an augmentation module that fertilizes the training set with generated samples to alleviate the scarcity of anomalous samples.
    \item \textbf{Most positive query} selects the top-k samples ordered by their anomaly scores.
    \item \textbf{Positive diverse query} combines anomaly scores with distance-based diversification into querying. 
    \item \textbf{Diverse query} \cite{li2023deep} utilizes k-means++ to initialize diverse clusters. The probability of another query from the unlabeled set is proportional to its distance to the closest sample already in the query set.
\end{itemize}

We implement GKDE~\footnote{\url{https://github.com/zxj32/uncertainty-GNN}}, OODGAT~\footnote{\url{https://github.com/SongYYYY/KDD22-OODGAT}}, BWGNN~\footnote{\url{https://github.com/squareRoot3/Rethinking-Anomaly-Detection}} and DAGAD~\footnote{\url{https://github.com/FanzhenLiu/DAGAD}} using the code published by their authors.
For FdGars and GeniePath, we use the code provided by an open-source library \footnote{\url{https://github.com/safe-graph/DGFraud}} for graph anomaly detection.
Regarding other active query strategies, we adopt the code~\footnote{\url{https://github.com/aodongli/Active-SOEL}} provided by \cite{li2023deep}, and use the same network structure as the encoder and anomaly score predictor in \method.

\subsection{Implementation Notes} \label{app:implementation}
We conduct all experiments on a Linux server with 64G RAM, and 2 NVIDIA GeForce RTX 2080TI with 11GB GPU memory.
We implement \method with Python 3.8.1 and PyTorch 2.0.1. 
We report our hyperparameter settings that are tuned on the validation set with 5 runs in \autoref{tab:hyperparameters}.

\begin{table}[h]
\caption{Hyperparameter settings of \method.}
\label{tab:hyperparameters}
\begin{tabular}{c|cccc}
\hline
              & Cora   & Citeseer & BlogCatalog & Flickr   \\ \hline
$\tau$        & 0.95   &   0.90   &   0.98      &   0.98   \\
$\alpha$      & 1.25   &   0.50   &   1.25      &   1.25   \\
$\beta$       & 0.50   &   2.00   &   1.00      &   0.50   \\
$\phi$        & 1.25   &   2.00   &   1.00      &   0.50   \\
$m$           & 24     &   24     &   18        &   27     \\ \hline
\end{tabular}
\end{table}
\end{document}